\documentclass{article}

\usepackage[preprint]{neurips_2025}

\usepackage[table,dvipsnames]{xcolor}

\usepackage{pifont}   

\usepackage{neurips_2025} 

\usepackage{hyperref}
\usepackage{pgfplots}
\usepackage{wrapfig}
\pgfplotsset{compat=1.18}  
\usepackage{tikz}
\usepackage[many]{tcolorbox}
\usepackage{amsmath}
\usepackage{multirow}
\usepackage[capitalize,noabbrev]{cleveref}
\usepackage{subcaption}
\usepackage{tcolorbox}
\tcbuselibrary{raster,skins}

\usepackage{microtype}
\usepackage{graphicx}
\usepackage{booktabs} 

\usepackage{amssymb}
\usepackage{mathtools}
\usepackage{amsthm}
\usepackage{xspace} 
\usepackage{mdframed}

\usepackage[textsize=tiny]{todonotes}

\theoremstyle{plain}

\theoremstyle{definition}
\newtheorem{definition}{Definition}

\theoremstyle{remark}

\newcommand{\sys}{\textsc{TokenSwap}\xspace}
\newcommand{\pmain}{\mathbf{p}^{\text{main}}}       
\newcommand{\paux}{\mathbf{p}^{\text{aux}}}         
\newcommand{\pfinal}{\mathbf{p}^{\text{final}}}   
\newcommand{\gramset}{\mathcal{G}}

\mdfdefinestyle{exampleframe}{%
    linecolor=gray!30,
    linewidth=0.5pt,
    backgroundcolor=gray!5,
    roundcorner=2pt,
    innertopmargin=8pt,
    innerbottommargin=8pt,
    innerleftmargin=8pt,
    innerrightmargin=8pt,
    skipabove=\baselineskip,
    skipbelow=\baselineskip
}

\definecolor{prefixcolor}{RGB}{0,0,0}
\definecolor{standardcolor}{RGB}{100,100,100}
\definecolor{tokenswapcolor}{RGB}{0,90,180}
\definecolor{pastelred}{RGB}{255,150,150}
\definecolor{pastelgreen}{RGB}{150,255,150}
\definecolor{pastelyellow}{RGB}{255,255,150}
\newcommand{\generationexample}[4]{%
    \begin{mdframed}[style=exampleframe]
    \end{mdframed}
    }

\usepackage[utf8]{inputenc} 
\usepackage[T1]{fontenc}    

\usepackage{url}            

\usepackage{amsfonts}       
\usepackage{nicefrac}       

\usepackage{algorithm}
\usepackage{algorithmic} 
\usepackage{colortbl}

\newcommand{\cmark}{\textcolor{black}{\ding{51}}} 
\newcommand{\xmark}{\textcolor{black}{\ding{55}}} 
\definecolor{pastelgreen}{RGB}{198,239,206}  
\definecolor{pastelyellow}{RGB}{255,255,153} 
\definecolor{pastelred}{RGB}{255,183,153}    

\definecolor{darkpurple}{RGB}{48, 25, 52}
\definecolor{darkbrick}{RGB}{128, 0, 0}  
\definecolor{darkforest}{RGB}{0, 59, 0}
\definecolor{darknavy}{RGB}{0, 0, 89}

\title{\sys: A Lightweight Method to Disrupt Memorized Sequences in LLMs}

\author{%
  Parjanya Prajakta Prashant~\thanks{Equal Contribution; Correspondence to Parjanya Prajakta Prashant <pprashant@ucsd.edu>, Kaustubh Ponkshe <ponkshekaustubh11@gmail.com>}\\
UC San Diego
  \And
  Kaustubh Ponkshe \(^*\)\\
  MBZUAI
  \And
  Babak Salimi\\
  UC San Diego
}

\begin{document}

\maketitle

\begin{abstract}

As language models scale, their performance improves dramatically across a wide range of tasks, but so does their tendency to memorize and regurgitate parts of their training data verbatim. This tradeoff poses serious legal, ethical, and safety concerns, especially in real-world deployments. Existing mitigation techniques, such as differential privacy or model unlearning, often require retraining or access to internal weights making them impractical for most users. In this work, we introduce \textsc{TokenSwap}, a lightweight, post-hoc defense designed for realistic settings where the user can only access token-level outputs. Our key insight is that while large models are necessary for high task performance, small models (e.g., DistilGPT-2) are often sufficient to assign fluent, grammatically plausible probabilities to common function words - and crucially, they memorize far less. By selectively swapping token probabilities between models, \textsc{TokenSwap} preserves the capabilities of large models while reducing their propensity for verbatim reproduction. Evaluations on Pythia-6.9B and Llama-3-8B show up to a 10$\times$ drop in exact memorization with negligible task degradation. Our method offers a practical, accessible solution for mitigating memorized generation in deployed LLMs.

\end{abstract}

\section{Introduction}
\label{sec:introduction}
Large language models (LLMs) such as \textsc{GPT--4}, \textsc{Gemini}, and \textsc{Llama} have demonstrated strong performance across a wide range of tasks, from natural language understanding to complex reasoning~\citep{achiam2023gpt,team2023gemini,dubey2024llama}. These capabilities are driven by their massive parameter counts and extensive training corpora, enabling human-level fluency and impressive reasoning across domains. Often referred to as emergent properties, such abilities arise directly from scale, with well-established scaling laws predicting performance gains. However, increased scale also introduces a critical drawback: the tendency of LLMs to memorize and reproduce parts of their training data.~\citep{carlini2021extracting, carlini2022quantifying, biderman2024emergent, nasr2025scalable}

One of the most pressing consequences of memorization is the
verbatim or near-verbatim generation of training data \citep{karamolegkou2023copyright,tirumala2022memorization,allen-zhu2025physics}. Although memorization is an inherent property and not necessarily harmful, its consequence of verbatim generation leads to plagiarism and copyright violation. This behavior poses serious risks to both model providers and end-users. Providers may face legal challenges, including copyright infringement lawsuits~\citep{karamolegkou2023copyright, grynbaum2023times, panwar2025generative}, while users unknowingly risk legal liability by reproducing protected content. Crucially, the threat is not limited to exact substring matches: even approximate or near-verbatim outputs can constitute infringement, as evidenced by lawsuits like the \textit{New York Times} case against OpenAI for near-verbatim content generation~\citep{freeman2024exploring}.

\begin{figure*}[h]
    \centering
    \includegraphics[width=\textwidth]{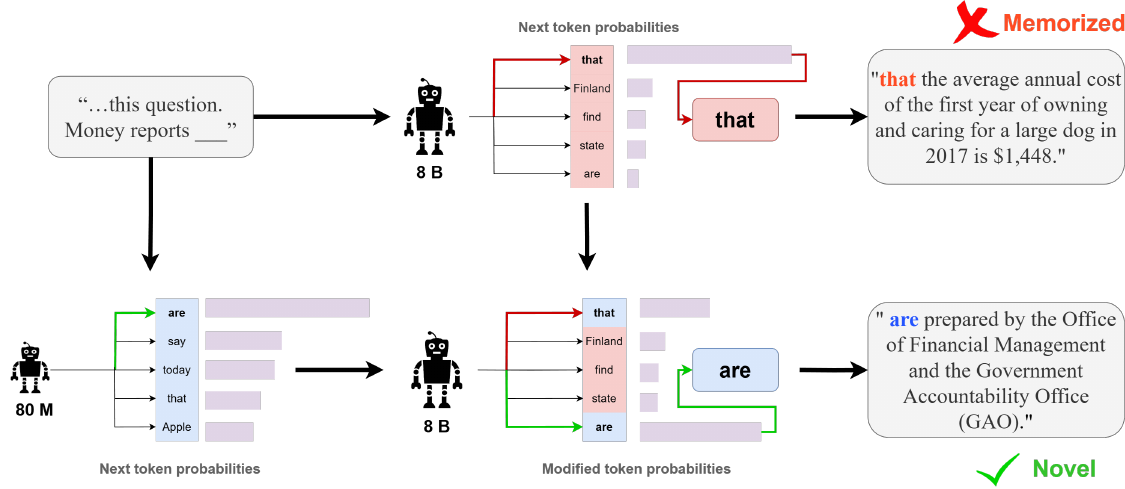}
    \caption{Overview of \textsc{TokenSwap}. Our approach replaces token probabilities of high-frequency "grammar-based" tokens with those from a small auxiliary language model. This mitigates memorized generation while maintaining fluency and model performance. The top path shows standard LLM generation, while the bottom path demonstrates how \textsc{TokenSwap} alters token selection to disrupt memorization and produce novel text.}
    \label{fig:tkswap_overview}
\end{figure*}

We consider the perspective of a typical user of commerical LLMs such as GPT-4~\citep{achiam2023gpt}, Gemini~\citep{team2023gemini}, Llama3~\citep{dubey2024llama}, and Deepseek~\citep{liu2024deepseek}. These models do not share their training data and many do not make their weights publicly available. Even in cases where weights are openly shared, hosting a production-grade LLM requires substantial memory resources, rendering it impractical for the average user. Consequently, it is reasonable to assume that most users can only interact with these models through APIs hosted on external servers, with access limited to model outputs such as token-level logits. Despite these practical constraints, to our knowledge, none of the existing methods, whether designed to prevent memorization or mitigate verbatim output, can effectively operate under such limited access conditions.

\textbf{Existing methods require access to training data and/or model weights} Approaches to address memorization are broadly categorized into pre-training and post-training interventions. Pre-training methods include deduplication \citep{kandpal2022deduplicating}, differential privacy (DP) \citep{abadi2016deep}, and selective token exclusion during training \citep{hans2024like}. While these approaches can reduce memorization, they often incur substantial computational costs and degrade model performance \citep{Anil2021}. Post-training interventions focus on unlearning techniques that attempt to modify specific neurons and weights or utilize finetuning methods to prevent models from generating memorized content~\citep{maini2023can, sakarvadia2024mitigating, chen2025parapo, russinovich2025obliviate}. However, these methods remain susceptible to training data extraction~\citep{shumailov2024ununlearning}, often impair general model capabilities~\citep{huang2024demystifyingverbatimmemorizationlarge}, and can lead to unintended forgetting of critical aspects such as safety guardrails~\citep{wang2025more}. This challenge is further complicated by theoretical findings suggesting that some degree of memorization may be inherent to achieving generalization in learning algorithms \citep{attias2024information}.

In contrast, another line of work focuses on preventing the generation of memorized content at inference time without modifying model weights. These approaches include blocking exact matches to training data \citep{ippolito2022preventing} or combining logits from multiple models trained on disjoint datasets \citep{abad2024copyright}. However, these methods too require access to training data or multiple LLMs trained on strictly disjoint datasets. Table~\ref{tab:methods-comparison} summarizes the various approaches and assumptions under which they operate (see Appendix~\ref{appendix:related-work} for a comprehensive review).

\textbf{Memorization scales with size} The propensity to reproduce training data consistently increases with the size of the language model~\citep{carlini2022quantifying, biderman2024emergent}. Since model performance generally scales positively with size, users are forced into a trade-off between obtaining high performance and mitigating memorized generations. Figure~\ref{fig:memorizationvsperformance} demonstrates this relationship using a series of Pythia models, showing the trade-off between memorization and cross-entropy loss. 

\begin{wrapfigure}[18]{r}{0.5\textwidth}
\vspace{-1em}
\begin{tikzpicture}
\begin{axis}[
    width=7.2cm,
    height=5cm,
    xlabel=Cross Entropy Loss (CE Loss) $\downarrow$,
    ylabel=Exact Match Rate (\%) (EMR) $\downarrow$,
    xmin=2.5,
    xmax=4.5,
    ymin=-5,
    ymax=70,
    xtick={2.5,3.0,3.5,4.0,4.5},
    ytick={0,20,40,60},
    xlabel near ticks,
    ylabel near ticks,
    axis lines=box,
    clip=true,
    scatter/classes={
        normal={mark=diamond*,draw=none,fill=orange,draw opacity = 0},
        sys={mark=*,draw=none,fill=red!60!white,draw opacity = 0}
    },
    tick style={draw=none},
    every axis label/.style={font=\small},
    every tick label/.style={font=\small},
    legend pos=north east,
    legend style={font=\small},
    axis background/.style={fill=gray!15},
    xmajorgrids=true,
    ymajorgrids=true,
    grid style={white, line width=.5pt},
    xticklabel style={opacity=1},
    yticklabel style={opacity=1},
    extra x ticks={2.5,3.0,3.5,4.0,4.5},
    extra y ticks={0,20,40,60},
    extra tick style={grid=none, tick style={draw=none}},
]
\addplot[scatter,only marks,
    scatter src=explicit symbolic,
    mark size=3pt,    
    mark options={draw=none}]
table[meta=label] {
x    y      label
2.78 65.22  normal
2.85 45.11  normal
2.98 36.96  normal
3.06 32.61  normal
3.24 16.30  normal
3.61 5.98   normal
4.05 1.09   normal
};

\addplot[scatter,only marks,
    scatter src=explicit symbolic,
    mark size=2.5pt]
table[meta=label] {
x    y      label
2.87 8.7    sys
2.95 7.61   sys
3.07 5.43   sys
3.15 5.43   sys
3.30 5.98   sys
3.65 3.80   sys
4.05 1.09   sys
};

\addplot[orange,dotted,thick] coordinates {
    (2.78, 65.22)
    (2.85, 45.11)
    (2.98, 36.96)
    (3.06, 32.61)
    (3.24, 16.30)
    (3.61, 5.98)
    (4.05, 1.09)
};

\addplot[red!60!white,solid,thick] coordinates {
    (2.87, 8.7)
    (2.95, 7.61)
    (3.07, 5.43)
    (3.15, 5.43)
    (3.30, 5.98)
    (3.65, 3.80)
    (4.05, 1.09)
};

\node[anchor=west, font=\tiny, text=black] at (axis cs:2.78,65.22) {6.9B};
\node[anchor=west, font=\tiny, text=black] at (axis cs:2.85,45.11) {2.8B};
\node[anchor=west, font=\tiny, text=black] at (axis cs:2.98,36.96) {1.4B};
\node[anchor=west, font=\tiny, text=black] at (axis cs:3.06,32.61) {1B};
\node[anchor=west, font=\tiny, text=black] at (axis cs:3.24,16.30) {410M};
\node[anchor=west, font=\tiny, text=black] at (axis cs:3.61,5.98) {160M};
\node[anchor=west, font=\tiny, text=black] at (axis cs:4.05,1.09) {70M};

\node[anchor=east, font=\tiny, text=black] at (axis cs:2.87,8.7) {6.9B};
\node[anchor=north, font=\tiny, text=black] at (axis cs:2.95,7.61) {2.8B};
\node[anchor=north, font=\tiny, text=black] at (axis cs:3.07,5.43) {1.4B};
\node[anchor=south, font=\tiny, text=black] at (axis cs:3.15,5.43) {1B};
\node[anchor=north, font=\tiny, text=black] at (axis cs:3.30,5.98) {410M};
\node[anchor=north, font=\tiny, text=black] at (axis cs:3.65,3.80) {160M};
\node[anchor=east, font=\tiny, text=black] at (axis cs:4.05,1.09) {};

\legend{Standard, \sys}

\end{axis}
\end{tikzpicture}
\vspace{-2 em} 
\caption{\small
Memorization (EMR) vs Performance (CE Loss) across different model sizes. Larger, more capable models exhibit higher memorization. \sys, with Pythia-70M as the auxiliary model, achieves low memorization rates while maintaining competitive performance. Details in Section~\ref{subsection:wild} and Section~\ref{section:discussion}.}

\label{fig:memorizationvsperformance}
\end{wrapfigure}

In this work, we present \sys, an inference-time method that significantly alleviates this tradeoff by combining large model performance with small model memorization (Figure~\ref{fig:memorizationvsperformance}). \sys selectively replaces the probabilities of a subset of common grammar tokens (e.g., ``the'', ``of'', ``and'') of the large main model with those from a small auxiliary model. This technique disrupts the verbatim generation by breaking the high-probability paths that lead to verbatim reproduction. This disruption has a cascading effect: once one token deviates from the memorized sequence, all subsequent predictions are conditioned on this altered context, further preventing reproduction. Importantly, since small models reliably approximate probabilities for common grammatical tokens, \sys preserves the large model's performance. For auxiliary models of size much smaller than the main model, this provides a verbatim memorization mitigation method which requires access \textit{neither to the training data nor the model weights}. Since we treat the effect of memorization, and not the cause itself, our method is able to reduce verbatim generation at inference time.  

We extensively evaluate \sys through both controlled experiments and real-world deployments. In controlled fine-tuning experiments (Section~\ref{subsection:finetuning}), \sys achieves a 50-800$\times$ reduction in verbatim generation compared to undefended models. Evaluations on commercial-grade models such as Pythia-6.9b and Llama-3-8b (Section~\ref{subsection:wild}) demonstrate reductions in verbatim generation by upto 10$\times$, without compromising downstream task performance. Furthermore, comparisons with Goldfish~\citep{hans2024like} show that \sys matches or surpasses the effectiveness of state-of-the-art pre-training methods (Section~\ref{subsection:goldfish}).

\begin{table}[ht]
\centering
\caption{Comparison of \sys with existing methods based on their assumptions. \sys uniquely avoids requiring access to model weights or the copyrighted training corpus. While it employs an auxiliary model, the memory overhead is small \(\approx 1\%\) due to the small size of the auxiliary model. \textcolor{blue}{PT}: pre-training, \textcolor{orange}{UL}: unlearning, \textcolor{green!40!black}{FT}: fine-tuning, \textcolor{violet}{Inf}: inference-time.}

\setlength{\tabcolsep}{5pt}
\small
\begin{tabular}{lccc}
\toprule
 & Model Access & Copyrighted Corpus Access & Inference Overhead \\ 
\midrule
Deduplication [\citenum{kandpal2022deduplicating}] \textcolor{blue}{PT}        & \cellcolor{pastelred}weights & \cellcolor{pastelred}\cmark & \cellcolor{pastelgreen}\xmark \\ 
Goldfish [\citenum{hans2024like}] \textcolor{blue}{PT}                          & \cellcolor{pastelred}weights & \cellcolor{pastelred}\cmark & \cellcolor{pastelgreen}\xmark\\ 
Balanced subnet [\citenum{sakarvadia2024mitigating}] \textcolor{orange}{UL}    & \cellcolor{pastelred}weights & \cellcolor{pastelred}\cmark & \cellcolor{pastelgreen}\xmark \\ 
Obliviate [\citenum{russinovich2025obliviate}] \textcolor{green!40!black}{FT}  & \cellcolor{pastelred}weights & \cellcolor{pastelred}\cmark & \cellcolor{pastelgreen}\xmark \\ 
MemFree [\citenum{ippolito2022preventing}] \textcolor{violet}{Inf}             & \cellcolor{pastelgreen}logits  & \cellcolor{pastelred}\cmark & \cellcolor{pastelred}querying entire dataset \\
CP-Fuse [\citenum{abad2024copyright}] \textcolor{violet}{Inf}                  & \cellcolor{pastelgreen}logits  & \cellcolor{pastelred}\cmark & \cellcolor{pastelred}twice of standard generation\\
\sys\ \textcolor{violet}{Inf}                                              & \cellcolor{pastelgreen}logits  & \cellcolor{pastelgreen}\xmark  & \cellcolor{pastelyellow}small auxiliary model\\
\bottomrule
\end{tabular}
\label{tab:methods-comparison}
\end{table}

\section{Preliminaries}
\label{section:preliminaries}
\subsection{Language Models: Notation and Setup}

We consider auto-regressive language models that model the log-probability of a token conditioned on all previous tokens in a sequence. They operate over a vocabulary \(\mathcal{V} = \{v_1, \ldots, v_{|\mathcal{V}|}\}\) of typically \(|\mathcal{V}| \approx 10^5-10^6\) tokens. Given an input prompt \((x_{-l_p}, \ldots, x_{-1}) \in \mathcal{V}^{l_p}\) of length \(l_p\) followed by a response sequence \((x_0, \ldots, x_{l-1}) \in \mathcal{V}^{l}\) of length \(l\), an auto-regressive language model parametrizes the joint probability by factorizing over conditional probabilities:

\begin{equation}
    p(x_0, \ldots, x_l | x_{-l_p}, \ldots, x_{-1}) \;=\; \prod_{i=0}^l p(x_i \,|\, x_{<i}),
\label{eq:autoregressive}
\end{equation}
For each position \(i\), the model outputs a distribution \(\mathbf{p}_i[v]\) over \(\mathcal{V}\), where \(\mathbf{p}_i[v] = p(x_i = v | x_{<i})\).

Since language models are trained to maximize the likelihood of observed sequences, they tend to assign high probabilities to tokens that frequently follow specific prefixes during training. This increases the risk of memorization and \emph{verbatim reproduction} of training data.

\subsection{Extractable Memorization}
Memorization in language models can manifest in various ways, but a practically relevant and widely adopted framework is \emph{extractable memorization}~\citep{carlini2021extracting, carlini2022quantifying}. \citet{carlini2021extracting} demonstrate that models can be induced to regurgitate training sequences when prompted with prefixes from their training data. The following definition formalizes this concept:

\begin{definition}[Extractable Memorization]
\label{definition:memorization}
A sequence \(x = (x_0, \ldots, x_{l-1})\) of length \(l\) is considered \emph{extractable with \(l_p\) tokens of context} from a language model \(p\) if there exists a prefix \(x_{-} = (x_{-l_p}, \ldots, x_{-1})\) of length \(l_p\) such that 
\([\,x_{-} \,\|\, x\,]\) appears in the training data of \(p\), and \(p\) reproduces \(x\) via greedy decoding. 
\\
Formally, for each \(i \in \{\,0,\ldots,l-1\}\):
\[
x_i \;=\;
\arg\max_{x' \in \mathcal{V}} \; p\bigl(x' \,|\, x_{<i},x_{-}\bigr).
\]
\end{definition}

This definition is practically useful because: (1) it aligns with real-world risks of copyright and memorized generation~\citep{nasr2311scalable, karamolegkou2023copyright}, (2) it provides a concrete, testable condition that can be evaluated on real models, and (3) it extends to models of different sizes, capturing the well-documented trend that larger models memorize more data~\citep{carlini2022quantifying, biderman2024emergent}. This scaling behavior is important in motivating our methodology in Section~\ref{section:methodology}.

\section{Methodology}
\label{section:methodology}

As discussed earlier, small language models (e.g., DistilGPT-2, Pythia-70M) have lower propensity to reproduce training data compared to large models (e.g., Llama3, GPT-4). We introduce \sys, a lightweight, post-hoc method that combines the strengths of both model scales: large-model performance with small-model memorization. During inference, \sys replaces the probabilities for selected tokens of a large model with those a small model. 

\paragraph{Algorithm}
\label{subsec:proposed_algorithm}
Let \(\pmain\) and \(\paux\) denote the probability distributions of the main and auxiliary models respectively, where \(\pmain(x_t \mid x_{<t})\) and \(\paux(x_t \mid x_{<t})\) represent their token probabilities conditioned on previous tokens. We assume the parameter count of the main model significantly exceeds that of the auxiliary model. Given these models, \sys selectively replaces probabilities for a fixed subset of tokens \(\gramset \subset \mathcal{V}\). The complete procedure is formalized in Algorithm \ref{alg:tokenswap}.

At each position \(i\), \sys\ queries both \(\pmain\) and \(\paux\) to obtain probability distributions conditioned on the current context \(x_{<i}\). For tokens in subset \(\gramset \subset \mathcal{V}\), probabilities from the main model are replaced with scaled probabilities from the auxiliary model, with scaling factor \(\alpha\) ensuring the final distribution \(\pfinal\) remains a valid distribution. This prevents reproduction of memorized sequences: if any token \(x_i\) in a memorized sequence belongs to \(\gramset\), its probability under \(\pfinal\) is determined by the auxiliary model. Since the auxiliary model memorizes less, this disrupts the chain of conditional probabilities required for verbatim generation of most sequences. Importantly, for tokens \(v \notin \gramset\), their probabilities remain unchanged, i.e., \(\pfinal_i[v] = \pmain_i[v]\).
\begin{algorithm}[!ht]
\caption{\sys}
\label{alg:tokenswap}
\begin{algorithmic}[1]
\REQUIRE Main model \(\pmain\), auxiliary model \(\paux\), token subset \(\gramset\), prompt \(x_{<0}\)
\FOR{\(i = 0, 1, \ldots\)}
    \STATE \(\pmain_i \gets \pmain(\cdot | x_{<i})\) \COMMENT{Get main model probabilities}
    \STATE \(\paux_i \gets \paux(\cdot | x_{<i})\) \COMMENT{Get auxiliary model probabilities}
    \STATE \(\alpha \gets \frac{\sum_{v \in \gramset} \pmain_i[v]}{\sum_{v \in \gramset} \paux_i[v]}\) \COMMENT{Compute normalization}
    \FOR{\(v \in \mathcal{V}\)}
        \STATE \(\pfinal_i[v] \gets \begin{cases}
            \pmain_i[v], & \text{if } v \notin \gramset \\
            \alpha \cdot \paux_i[v], & \text{if } v \in \gramset
        \end{cases}\)
    \ENDFOR
    \STATE \(x_i \sim \pfinal_i\) \COMMENT{Sample next token}
\ENDFOR
\end{algorithmic}
\end{algorithm}

\paragraph{Selecting \(\gramset\) for Effective Memorization Disruption}  
The choice of \(\gramset\) affects both memorization and model performance. By modifying token probabilities, \sys disrupts memorized sequences while preserving fluency. However, not all tokens are equally effective for this purpose. \(\gramset\) should consist of tokens that frequently appear in memorized text, as replacing their probabilities reduces the likelihood of exact reproduction. At the same time, modifying inappropriate tokens can degrade model performance, especially for specialized tasks. For instance, if \(\gramset\) includes numeric tokens, mathematical reasoning may degrade. Therefore, \(\gramset\) should satisfy two key criteria. First, it must contain frequently occurring tokens. Second, it should avoid tokens where probability replacement impacts the model's capabilities.

Empirical studies suggest that small models correctly approximate the probabilities of high-frequency function words while diverging more on rare or domain-specific terms~\citep{pinto2024fair}. Additionally, small language models (\(\approx 100M\)) can generate coherent and grammatically correct text~\citep{Eldan2023}. Based on these insights, we construct \(\gramset\) from grammar-based high-frequency tokens (e.g. - 'the', 'in'). Further, since \(\gramset\) consists of high-frequency words, there exists a natural one-to-one mapping between tokens even when \(\pmain\) and \(\paux\) use different tokenizers and vocabularies. While this approach is well-suited for natural language, structured domains such as code may require domain-specific adaptations. Additional details on the construction of \(\gramset\) are provided in Appendix~\ref{appendix:construction_grammar} and~\ref{appendix:listofgramset}.

\section{Experiments}
\label{section:experiments}
In this section, we demonstrate the effectiveness of \textsc{TokenSwap}, in both controlled and real-world settings. Our experiments evaluate \sys along two dimensions:
\begin{itemize}
    \item The method's efficacy in preventing exact \textit{and approximate} reproduction of training data.
    \item The impact on model performance across common-sense reasoning, language and fluency.
\end{itemize}

We evaluate \sys across three settings to demonstrate its effectiveness. In Section~\ref{subsection:finetuning}, we deliberately induce memorization through extensive fine-tuning on small datasets to stress-test our defense. Section~\ref{subsection:wild} evaluates \sys on production-grade models including Pythia-6.9B and Llama-3-8B. Finally, in Section~\ref{subsection:goldfish}, we compare against Goldfish~\citep{hans2024like}, a pre-training method specifically designed to reduce memorization, showing that our post-hoc approach achieves comparable results without requiring model retraining.
\subsection{Extreme Memorization}
\label{subsection:finetuning}

In order to rigorously evaluate \sys, we create an extreme test case by deliberately inducing memorization through extensive fine-tuning. While \sys can be applied to real-world models directly, our baselines require specific experimental conditions for comparison. Similar extreme test cases have been generated to evaluate memorization in prior work~\citep{hans2024like, abad2024copyright}. Following \citet{abad2024copyright}, we fine-tune a Llama-3.2-3B model \citep{dubey2024llama} on 2,000-sequence subsets from two datasets: MathAbstracts~\citep{zhang2024autonomous} and WritingStories~\citep{fan2018hierarchical}. We train for 50 epochs to deliberately amplify memorization beyond typical levels.

\paragraph{Memorization Metrics}
Our analysis employs both exact and approximate memorization and performance metrics to ensure a comprehensive assessment. Exact memorization is measured through \textit{Matching Length (ML)}, which the number of verbatim characters or tokens generated before first deviation, and \textit{Exact Matching Rate (EMR)}, which computes the fraction of sequences reproduced verbatim. To capture partial memorization, we use the \textit{ROUGE-L} score, which identifies the longest common non-contiguous subsequence and  gives a score between 0 and 1, and the \textit{Normalized Levenshtein Distance}, which quantifies the minimum number of edits needed to transform generated text into the original sequence. Lower scores indicate reduced memorization for Matching Length, Exact Matching Rate, and ROUGE-L. Higher scores are better for Normalized Levenshtein Distance. These metrics are widely used to evaluate verbatim and approximately verbatim generation~\citep{karamolegkou2023copyright, hans2024like, abad2024copyright}.

\paragraph{Performance Metrics} Since our setup intentionally induces extreme memorization, standard performance metrics are not meaningful. Nonetheless, we report cross-entropy loss on a held-out validation set in Appendix~\ref{appendix:cross-entropy-extreme-memorization}. 

\paragraph{Setup and Inference-time Baselines}
We compare against the two inference-time baselines: CP-Fuse~\citep{abad2024copyright}, which samples from weighted combinations of models trained on disjoint datasets, and MemFree~\citep{ippolito2022preventing}, which blocks exact $n$-gram matches to the training data. Standard refers to greedy decoding without any memorization mitigation. Both baselines rely on unrealistic assumptions—MemFree requires access to the training data, while CP-Fuse assumes access to two separately trained models on disjoint corpora. To assess CP-Fuse under more realistic conditions, we evaluate two variants: \textsc{CP-Fuse Half}, with perfectly disjoint sets of 1,000 sequences each, and \textsc{CP-Fuse Mixture}, with 1,500 sequences per model and 500 overlapping. For \sys, we employ DistilGPT-2 (80M)~\citep{distilgpt2} as \(\paux\). We construct \(\gramset\) with \(|\gramset| = 110\) tokens using high-frequency 'grammar-based' words. Additional details on \(\gramset\) are provided in Appendix~\ref{appendix:construction_grammar}.
For all experiments and methods, a prefix of 20 tokens is used and the next 128 tokens are greedily sampled (temperature = 0.0).

\begin{table}[!ht]
\small
\setlength{\tabcolsep}{4pt} 
\centering
\caption{\small Memorization for WritingPrompts and MathAbstracts datasets. ML: Matching Length, EMR: Exact Match Rate, Lev.: Normalized Levenshtein Distance}
\label{tab:memorization_finetuning}
\begin{tabular}{l|cccc|cccc}
\toprule
& \multicolumn{4}{c|}{WritingPrompts} & \multicolumn{4}{c}{MathAbstracts} \\
\cmidrule(lr){2-5} \cmidrule(lr){6-9}
Method & ML\(\downarrow\) & ROUGE-L\(\downarrow\) & Lev.\(\uparrow\) & EMR\(\downarrow\) & ML\(\downarrow\) & ROUGE-L\(\downarrow\) & Lev.\(\uparrow\) & EMR\(\downarrow\)\\
\midrule
Standard       & 464.0 & 0.89 & 0.10 & 83.4 & 450.4 & 0.98 & 0.03 & 93.6 \\
MemFree        & 17.4  & 0.29 & 0.63 & 0.0  & 6.7   & 0.44 & 0.55 & 0.0 \\
CP-Fuse-mix    & 280.3 & 0.58 & 0.37 & 49.2 & 233.7 & 0.62 & 0.36 & 47.1 \\
CP-Fuse-half   & 12.5  & 0.17 & 0.73 & 0.0  & 15.3  & 0.26 & 0.71 & 0.1 \\
\sys           & \textbf{19.7}  & \textbf{0.19} & \textbf{0.71} & \textbf{0.1}  & \textbf{53.0}  & \textbf{0.38} & \textbf{0.60} & \textbf{1.8} \\
\bottomrule
\end{tabular}
\end{table}

\paragraph{Results}
Table~\ref{tab:memorization_finetuning} demonstrates \sys's effectiveness in reducing memorization across both datasets. For WritingPrompts, \sys reduces EMR by 800x (from 83.4\% to 0.1\%) and ROUGE-L by 4.6x (from 0.89 to 0.19) compared to standard generation. On MathAbstracts, EMR decreases by 50x (from 93.6\% to 1.8\%) and ROUGE-L by 2.6x (from 0.98 to 0.38). CP-Fuse-half achieves slightly better results but requires disjoint training sets, while CP-Fuse-mix performs significantly worse due to dataset overlap. MemFree achieves the lowest scores on the exact memorization metrics (Exact Matching Rate and Matching Length) but performs poorly on approximate memorization metrics (ROUGE-L and Levenshtein). This shows that, while MemFree prevents verbatim generation, it still allows high levels of near-verbatim generation. The performance gap between WritingPrompts (EMR: 0.1\%) and MathAbstracts (EMR: 1.8\%) aligns with our intuition - \(\gramset\) was designed focusing on natural language tasks. Nevertheless, \sys achieves substantial memorization reduction for both domains. To complement our quantitative results, \textit{we provide qualitative examples of generations from the WritingPrompts dataset} in Figure~\ref{fig:memorization_comparison} and Appendix~\ref{appendix:examples}.

\newtcolorbox{prefixbox}{
  enhanced,
  colback=headercolor,
  colframe=headercolor,
  fonttitle=\bfseries\color{white},
  coltext=white,
  arc=2mm,
  boxrule=0.5pt,
  left=3mm,
  right=3mm,
  top=2mm,
  bottom=2mm,
  width=\textwidth,
}

\newtcolorbox{contentbox}[1][]{
  enhanced,
  colback=boxcolor,
  colframe=gray!50,
  fonttitle=\bfseries\color{white},
  colbacktitle=black,
  arc=2mm,
  boxrule=0.5pt,
  left=2mm,
  right=2mm,
  top=2mm,
  bottom=2mm,
  title=#1
}

\begin{figure}[!t]
\centering

\definecolor{boxcolor}{RGB}{255,240,220}
\definecolor{headercolor}{RGB}{25,25,112}

\begin{prefixbox}
\textbf{Prefix:} Magic – once a real and potent force but as the world population
\end{prefixbox}
\newlength{\threecolsep}
\setlength{\threecolsep}{1mm}
\begin{minipage}[t]{0.327\textwidth}
  \begin{contentbox}[Original Suffix]
    grew from millions to billions the shared mana per person is now negligible. A group of astronauts helplessly watching the Earth perish experience...
  \end{contentbox}
\end{minipage}\hspace{\threecolsep}%
\begin{minipage}[t]{0.327\textwidth}
  \begin{contentbox}[Standard Generation]
    {\color{red}grew from millions to billions the shared mana per person is now negligible. A group of astronauts helplessly watching the Earth perish experience... }
  \end{contentbox}
\end{minipage}\hspace{\threecolsep}%
\begin{minipage}[t]{0.327\textwidth}
  \begin{contentbox}[\sys Generation]
    {\color{red}grew} and the number of wizards and witches declined, the world began to suffer. Now the world suffers from a lack of magic, and the government is tasked with...
  \end{contentbox}
\end{minipage}

\caption{Comparison of text generation methods. Red text indicates memorized content. Standard generation reproduces the entire suffix verbatim, while \sys generates novel content.}
\label{fig:memorization_comparison}
\end{figure}

\subsection{Memorization in the wild}
\label{subsection:wild}
In this section, we demonstrate the efficacy of our approach on production-grade models. We assess the effectiveness of \sys on two pre-trained models: Pythia-6.9B~\citep{biderman2023pythiasuiteanalyzinglarge} and Llama-3-8B~\citep{dubey2024llama}. 

\paragraph{Pile-Memorized Dataset} For Pythia-6.9B, we evaluate on memorized sequences identified by \citet{chang2024localization} from the Pile dataset, consisting of 32-token prefixes and 48-token suffixes. After filtering to retain only natural language content (excluding code, URLs, etc.), we obtain 184 evaluation examples.

\paragraph{LeetCode Dataset} For Llama-3-8B, following previous work demonstrating LeetCode problem memorization~\citep{karamolegkou2023copyright}, we evaluate on 1,825 LeetCode problem statements~\citep{gzipchrist2021leetcode}. \textit{These problem statements are written in natural language}. Since the exact format of LeetCode problems in Llama's training data is unknown, we remove punctuation while calculating the memorization metrics. Additionally, instead of exact match rate, we use ROUGE-L \(> 0.8\)  as our threshold for identifying memorized content. Prefix length of 20 tokens is used and the next 100 tokens are sampled.

\paragraph{Evaluation Setup} 
We face two key limitations when comparing \sys with existing baselines. CP-Fuse requires models trained on disjoint datasets, but verifying this is difficult since most LLMs do not release training data. Even when available, disjoint datasets are unlikely given that most models train on overlapping web corpora like Common Crawl. Additionally, CP-Fuse requires identical tokenizers, limiting comparisons to models within the same family. Similarly, we cannot evaluate against MemFree due to unavailable training data (LLaMA) or prohibitively large datasets (Pythia uses the 800GB Pile~\citep{biderman2023pythiasuiteanalyzinglarge}). To ensure fair evaluation for CP-Fuse, we paired each model with a smaller counterpart: Pythia-2.8B with Pythia-6.9B, and Llama-3.2-3B with Llama-3-8B. Using smaller models actually favors CP-Fuse since they memorize less. We avoid very small models (\(<\)100M) as CP-Fuse needs roughly equally capable models (see Appendix~\ref{appendix:exp_cpfuse}). The setup for \sys follows Section~\ref{subsection:finetuning}. For LeetCode evaluation, we use both DistilGPT-2 and SmolLM-135M~\citep{allal2025smollm2} as auxiliary models. SmolLM is an instruction-tuned model, which enables evaluation on instruction-following tasks like MT-Bench where an instruct-capable auxiliary model is required. For memorization, we use the same metrics as Section~\ref{subsection:finetuning}.

\paragraph{Performance Metrics}We evaluate two key aspects: task performance and generation quality. For task performance, we assess five-shot learning on multiple commonsense reasoning benchmarks: BoolQ \citep{clark2019boolq}, SIQA \citep{sap2019socialiqa}, PIQA \citep{bisk2020piqa}, ARC-Challenge \citep{clark2018think}, ARC-Easy \citep{clark2018think}, OBQA \citep{mihaylov2018can}, and WinoGrande \citep{sakaguchi2021winogrande}. For generation quality, we report cross-entropy loss on samples from \textit{Slimpajama}~\citep{soboleva2023slimpajama}, which correlates with fluency~\citep{basu2020mirostat} and has been used to evaluate prior memorization mitigation work~\citep{hans2024like, abad2024copyright}. We also evaluate on MT-Bench~\citep{zheng2023judging}, which tests multi-turn conversation, instruction-following, and generation quality through realistic conversational scenarios. Note that MT-Bench and commonsense reasoning results are only reported for Llama-3-8B (LeetCode Dataset) since these require instruction-following capabilities not available in the base Pythia models.
\begin{table}[!ht]
\centering
\caption{\small Memorization metrics on LeetCode and Pile-Memorized datasets: ML: Matching Length, EMR: Exact Match Rate, Lev.: Normalized Levenshtein Distance  \&  Performance metrics: Cross Entropy Loss (CE Loss) on SlimPajama; MT-Bench with GPT-4 as a judge, Mean of scores on Commonsense Reasoning benchmarks. }
\label{tab:memorization_combined}
\small

\begin{subtable}{\textwidth}
\centering
\setlength{\tabcolsep}{2pt}
\begin{tabular}{l|cccc|c|c|c}
\toprule
\multicolumn{8}{c}{\textbf{LeetCode Dataset (Llama)}} \\
\midrule
Method & ML \(\downarrow\) & ROUGE-L \(\downarrow\) & Lev. \(\uparrow\) & ROUGE-L\(>0.8\) \(\downarrow\) & CE Loss  \(\downarrow\) &  MT-Bench \(\uparrow\) & Commonsense \(\uparrow\)\\
\midrule
Standard & 24.57 & 0.39 & 0.60 & 9.65 & 2.38 & 7.75 & 71.87  \\
CP-Fuse & 19.44 & 0.37 & 0.61 & 7.01 & 2.45 & 8.53 & 70.18\\
\sys\(^1\) & \textbf{6.04} & \textbf{0.27} & \textbf{0.71} & \textbf{0.96} & \textbf{2.52} & -& \textbf{71.87} \\
\sys\(^2\) & \textbf{8.58} & \textbf{0.30} & \textbf{0.69} & \textbf{1.92} & \textbf{2.43} & \textbf{7.78} & -\\
\bottomrule
\end{tabular}
\end{subtable}

\begin{subtable}{\textwidth}
\centering
\begin{tabular}{l|cccc|c}
\toprule
\multicolumn{6}{c}{\textbf{Pile-Memorized Dataset (Pythia)}} \\
\midrule
Method & ML \(\downarrow\) & ROUGE-L \(\downarrow\) & Lev. \(\uparrow\) & EMR \(\downarrow\) & CE Loss  \(\downarrow\) \\
\midrule
Standard & 151.6 & 0.80 & 0.18 & 65.22 & 2.80 \\
CP-Fuse  & 97.05  & 0.62 & 0.35 & 29.35 & 2.81 \\
\sys\(^1\) & \textbf{35.10} & \textbf{0.38} & \textbf{0.56} & \textbf{5.98} & \textbf{2.88} \\
\bottomrule
\end{tabular}
\end{subtable}

\footnotesize
$^1$DistilGPT-2 as auxiliary model. $^2$SmolLM-135M as auxiliary model.
\end{table}

\paragraph{Results} 
Table~\ref{tab:memorization_combined} demonstrates that \sys substantially reduces memorization across both datasets compared to standard generation and CP-Fuse. Exact match rate decreases by over 10x compared to standard generation and 5-7x compared to CP-Fuse on both datasets. The average matching length shows similar improvements, reducing by 4-5x versus standard and 3-4x versus CP-Fuse. The consistent improvements in approximate memorization metrics (ROUGE-L and Levenshtein distance) demonstrate that \sys robustly prevents verbatim generation rather than simply introducing small perturbations. CP-Fuse shows limited effectiveness in these real-world scenarios primarily because its core assumption of disjoint training datasets does not hold. Even when using different models, the inherent overlap in web-scale training corpora prevents CP-Fuse from effectively disrupting memorized sequences. 

\sys maintains task performance by selectively targeting only grammar-based tokens, leaving reasoning-critical content words unchanged. This preserves commonsense reasoning abilities, as shown by identical accuracy scores (71.87\%) compared to standard generation. The method also maintains fluency, evidenced by minimal cross-entropy increases and nearly equal MT-Bench scores. While CP-Fuse achieves better conversational performance (8.53 vs 7.78), it fails to verbatim generation, making it unsuitable for the desired goal.

\subsection{Comparison with Pre-training Methods}
\label{subsection:goldfish}
While previous sections demonstrate that \sys outperforms post-hoc baselines, we also compare with Goldfish~\citep{hans2024like}, a pre-training approach that reduces memorization by excluding a fraction $1/k$ of tokens from loss computation during training. Since pre-training large models using this loss is expensive, we evaluate on pre-trained goldfish models from \citet{hans2024like}. These models were trained on a subset of RedPajama~\citep{weber2024redpajama} combined with 2000 Wikipedia~\citep{bridge2001wikipedia} sequences. To induce memorization, the Wikipedia sequences were duplicated 50 times during training. We compare against models trained with $k \in \{3,4,32\}$. For \sys, we maintain the same experimental setup from Section~\ref{subsection:finetuning}. Following \citet{hans2024like}, we use identical prefix and suffix lengths for extraction of memorized sequences.
\begin{table*}[!ht]
\small
\caption{\small Comparison with Goldfish~\citep{hans2024like} for k \(\in\) \{3,4,32\}. }
\label{tab:goldfish}
\centering
\begin{tabular}{l|ccccc}
\toprule
Method & ML$\downarrow$ & ROUGE-L$\downarrow$ & Levenshtein$\uparrow$ & EMR$\downarrow$ & CE Loss$\downarrow$ \\
\midrule
Standard & 73.9 & 0.38 & 0.58 & 7.8 & 3.44 \\
Goldfish ($k$=3) & 12.7 & 0.23 & 0.72 & 0.0 & 3.54 \\
Goldfish ($k$=4) & 14.7 & 0.23 & 0.71 & 0.0 & 3.50 \\
Goldfish ($k$=32) & 58.1 & 0.35 & 0.60 & 2.5 & 3.44 \\
\sys & \textbf{12.4} & \textbf{0.22} & \textbf{0.72} & \textbf{0.1} & \textbf{3.44} \\
\sys + Goldfish ($k$=3)  & \textbf{7.9} & \textbf{0.21} & \textbf{0.73} & \textbf{0.0} & \textbf{3.57} \\
\end{tabular}
\end{table*}

\paragraph{Results} Table~\ref{tab:goldfish} shows \sys achieves comparable or superior performance to Goldfish across all memorization metrics. Notably, \sys obtains the best Matching Length, Rouge-L and Normalized Levenshtein distance scores while maintaining better cross-entropy than the Goldfish variants for \(k = 3,4\). The effectiveness of Goldfish varies with parameter $k$ - smaller values (more aggressive token exclusion) yield stronger memorization reduction but worse performance, as evidenced by higher cross-entropy. This illustrates a key advantage of \sys: we achieve similar memorization reduction without requiring modified training or reduced training data tokens. Figure~\ref{fig:rougelwithgoldfish} (Appendix~\ref{appendix:goldfish_plots}) further supports this finding, showing nearly identical ROUGE-L score distributions between \sys and Goldfish ($k$=3), indicating that our post-hoc approach matches the most aggressive pre-training variant. Furthermore, applying \sys to Goldfish ($k=3$) as the main model reduces memorization more than either method alone, demonstrating that our approach is orthogonal to pre-training methods and can enhance existing techniques.

\section{Discussion and Limitations}
\label{section:discussion}
In this section, we analyze \sys's behavior across different settings. We first perform ablations on the auxiliary model choice and the size of $\gramset$. We then analyze \sys across the Pythia model family to demonstrate significant improvements in the performance-memorization tradeoff. Finally, we discuss limitations and potential extensions of our method.

\paragraph{Choice of the Auxiliary Model} In Section~\ref{section:experiments} we test \sys with DistilGPT-2 as the auxiliary model. A natural question arises: \textit{What auxiliary model should one choose and how does the size of the auxiliary model affect memorized generation?} To answer this, we use the SmolLM family~\citep{allal2025smollm2} with three sizes (135M, 360M, 1.7B) and evaluate on both Pythia-6.9B (Pile-memorized dataset) and Llama-3-8B (LeetCode dataset). Detailed results are in Table~\ref{tab:app_memorization_combined} (Appendix~\ref{sec:appendix-ablation-aux}).

We observe a clear trend: smaller auxiliary models lead to less verbatim generation, confirming our hypothesis that \sys's effectiveness stems from low memorization in auxiliary models. Importantly, auxiliary model size has minimal impact on performance. MT-Bench scores show negligible variation across auxiliary models—this is particularly significant since MT-Bench evaluates overall sequence generation quality, unlike cross-entropy loss which measures token-level accuracy.
Therefore, any small model (\(\approx 100M\)) which can generate fluent text and predict grammar-based tokens well, such as DistilGPT-2 or SmolLM-135M, can be used effectively as an auxiliary model.

\paragraph{Ablations on \(\gramset\)} The subset of tokens \(\gramset\) is constructed by selecting grammar-based words from the top 500 most frequent English words, resulting in $|\gramset| = 110$ (details in Appendix~\ref{appendix:construction_grammar}). To understand the impact of \(\gramset\) size on memorization reduction, we ablate by constructing \(\gramset\) from the top $k$ most frequent words for $k \in \{10, 50, 100, 500, 2500\}$, yielding $|\gramset| \in \{9, 43, 66, 110, 136\}$. We evaluate on the Pile-memorized task using Pythia-6.9B as the main model and Pythia-70M as auxiliary (see Appendix~\ref{appendix:sizeofgramset} for complete results). We observe a clear trend: as $|\gramset|$ increases, memorization decreases significantly. For example, EMR drops from 22.28\% ($|\gramset| = 9$) to 8.15\% ($|\gramset| = 136$). This makes intuitive sense—larger \(\gramset\) enables the auxiliary model to disrupt memorized sequences more frequently. The cross-entropy loss remains largely stable, indicating minimal performance degradation with increase in \(|\gramset|\).

\paragraph{Performance-Memorization Tradeoff} We analyze how \sys affects the tradeoff between model performance and memorization across seven Pythia models (70M to 6.9B parameters), using Pythia-70M as the auxiliary model. Figure~\ref{fig:memorizationvsperformance} shows exact match rate (EMR) versus cross-entropy loss—lower values are better for both metrics. Standard generation faces a severe tradeoff: reducing memorization from 45\% to 6\% EMR costs 0.7 points in cross-entropy (2.85 → 3.55). \sys considerably improves this tradeoff. At similar performance levels (cross-entropy \(\approx\) 2.87), \sys achieves 8.7\% EMR versus 45\% for standard models—an 8× memorization reduction. Even when targeting very low memorization (6\% EMR), \sys maintains cross-entropy at 3.07, significantly outperforming standard models at equivalent memorization levels.

\paragraph{Limitations and Future work} One limitation of our work is that in the rare cases where the small auxiliary model memorizes a sequence, our approach will preserve that memorization. However, in practice, small auxiliary models (\(\approx\) 100M parameters) memorize very little, and we empirically match or outperform existing baselines without requiring access to training data or restrictive assumptions like disjoint datasets. Additionally, while pre-training or unlearning mitigation methods are impractical for large models, they can be applied to small models since these are often open-source with accessible training data. Therefore, we expect future development in small models with low memorization. This makes our work even more significant: \textit{any advance in pre-training or unlearning methods to reduce memorization in small models can be immediately extended to large models using \sys}. Second, our current implementation focuses on natural language tasks. A promising direction for future work is extending \sys to other domains such as code generation.

\section{Conclusion}\sys offers several key advantages for mitigating memorized generation in language models: it operates without requiring access to model weights or training data, and makes no assumptions about the underlying training distribution. Our experiments demonstrate 10-800× reductions in verbatim generation, matching or exceeding baselines that assume access to training data, disjoint models, or require pre-training their own models. Importantly, this comes at minimal cost to model performance. \sys maintains performance on commonsense reasoning tasks, and our MT-Bench evaluation shows that it preserves fluency, instruction-following, and conversational abilities. This makes \sys a practical solution for both providers and users of LLMs.

\newpage
\bibliography{ref}
\bibliographystyle{plainnat}

\newpage
\appendix
\onecolumn
\section{Related Work}
\label{appendix:related-work}

\paragraph{Memorization in LLMs}
LLMs have been shown to memorize and potentially reproduce copyrighted information from their training data \citep{carlini2021extracting, carlini2022quantifying, karamolegkou2023copyright, sok_mem}.  This is demonstrated through prefix attacks, where models prompted with training data prefixes generate their memorized completions. \citet{schwarzschild2024rethinking} formalize this notion based on adversarial compression, requiring that any memorized sequence must be longer than the prefix used to elicit it. \citet{zhou2023quantifyinganalyzingentitylevelmemorization} and  \citet{nasr2311scalable} demonstrate that large-scale training data can be extracted without access to training prefixes. Studies further indicate a correlation between model scale and memorization, with larger models regurgitating higher proportions of their training data \citep{carlini2022quantifying, zhou2023quantifyinganalyzingentitylevelmemorization, biderman2024emergent}.

\paragraph{Pre-training}

Several training-time strategies reduce memorization and verbatim generation, but often at the cost of accessibility or performance. De-duplication~\citep{kandpal2022deduplicating} is limited by pervasive near-duplicates in large-scale corpora. Differential Privacy (DP)~\citep{abadi2016deep} offers formal guarantees, but degrades performance and is computationally costly~\citep{Anil2021, elkin2023can}. Other methods such as token masking~\citep{hans2024like} and early stopping~\citep{mireshghallah2022memorization, pinto2024extracting} show some promise but remain expensive, degrade model performance and are unavailable to end users.

\paragraph{Unlearning and Finetuning}  
Post-training approaches offer alternative strategies to reduce memorization. Unlearning methods~\citep{maini2023can, jang2022knowledge, sakarvadia2024mitigating, chang24neuralsurgery} modify internal weights linked to memorized content. Others remove sequences via gradient ascent~\citep{barbulescu24unlearning}, steer activations away from memorization-correlated directions~\citep{suri25steering}, or fine-tune with losses discouraging verbatim recall~\citep{russinovich2025obliviate, chen2025parapo}. However, these methods require access to model internals and often degrade utility~\citep{huang2024demystifyingverbatimmemorizationlarge, suri25steering, chen2025parapo}.

\paragraph{Inference time} The two methods most relevant to our work are MemFree~\citep{ippolito2022preventing} and CP-Fuse~\citep{abad2024copyright}. These methods operate during generation and do not assume access to model internals. MemFree filters next-token outputs to block $n$-gram matches from the training set. It requires access to the full training corpus, often unavailable or prohibitively large for end users. Further, MemFree often degrades fluency by introducing unnatural punctuation~\citep{abad2024copyright}. CP-Fuse combines logits from two LLMs trained on disjoint corpora. This is rarely practical since most production-grade LLMs are trained on internet-scale data. Also, CP-Fuse requires the tokenizers of the two models to be the same. In contrast, our method can mitigate memorization in real-world models trained on internet-scale data.

\paragraph{Speculative decoding} Speculative decoding approaches accelerate inference by generating candidate tokens from a small draft model, which are selectively accepted by the large model~\citep{li22contrastive,  leviathan23speculative, chen23speculative, stern18blockwise, xia23speculative, kim23bild}. These methods preserve the model distribution and do not aim to mitigate verbatim generation. Further, if all candidates are rejected, the tokens are generated by the large model. In contrast, \sys modifies the large model’s distribution to reduce verbatim generation.

\section{Additional Experiments}
\label{sec:appendix-additional}
\subsection{Cross-Entropy for Extreme Memorization}
\label{appendix:cross-entropy-extreme-memorization}
\begin{table}[!h]
\centering
\caption{Validation Cross-entropy loss on WritingPrompts and MathAbstracts. Lower values \(\downarrow\) indicate better performance.}
\label{tab:performance_finetuning}
\begin{tabular}{lcc}
\toprule
Method & WritingPrompts & MathAbstracts \\
\midrule
Standard & 6.68 & 4.94 \\
MemFree & 6.68 & 4.94 \\
CP-Fuse-mixture & 9.38 & 6.89 \\
CP-Fuse-half & 9.43 & 6.67 \\
\sys & 5.98 & 4.65 \\
\bottomrule
\end{tabular}
\end{table}
Table~\ref{tab:performance_finetuning} reports the cross-entropy on a held-out validation set. \sys achieves the lowest cross-entropy loss across both datasets (5.98 and 4.65 for WritingPrompts and MathAbstracts respectively). The superior performance, even compared to standard generation, suggests our method effectively disrupts memorization pathways while preserving model capabilities. For sequences not in the training set, MemFree and Standard produce identical generations. Therefore, their cross-entropy values on a held-out validation set are the same.

\subsection{Commonsense Reasoning Results}\
Table~\ref{tab:combined_benchmark_results} reports performance across various commonsense reasoning benchmarks. \sys matches the performance of standard generation because our method does not affect token prediction for non-grammar tokens. This demonstrates that \sys achieves substantial memorization reduction without affecting task performance and reasoning.
\label{appendix:commonsense_reasoning}
\begin{table*}[!ht]
\setlength{\tabcolsep}{3pt}
\centering
\caption{Performance comparison on commonsense reasoning and general alignment benchmarks. All values are accuracy percentages or MT-Bench scores; higher is better (\(\uparrow\)).}
\label{tab:combined_benchmark_results}
\small
\begin{tabular}{l|cccccc}
\toprule
Method & WinoGrande \(\uparrow\) & PIQA \(\uparrow\) & OpenBookQA \(\uparrow\) & BoolQ \(\uparrow\) & ARC-E \(\uparrow\) & ARC-C \(\uparrow\)\\
\midrule
Standard & 54.69 & 64.84 & 76.56 & 70.31 & 82.03 & 82.81\\
CP-Fuse & 54.69 & 64.84 & 77.34 & 58.59 & 83.59 & 82.03  \\
\sys  & 54.69 & 64.84 & 76.56 & 70.31 & 82.03 & 82.81  \\
\bottomrule
\end{tabular}
\end{table*}

\subsection{Ablation on size of \(\gramset\)}
\label{appendix:sizeofgramset}
In this paper, \(\gramset\) is constructed by selecting grammar-based words from the top 500 most frequent English words, yielding 110 words in total (see Appendix~\ref{appendix:construction_grammar} for further details).

In this section, we ablate the size of \(\gramset\) by constructing it from the top $k$ most frequent English words for $k \in \{10, 50, 100, 500, 2500\}$. We evaluate on the Pile-memorized dataset using Pythia-6.9B as the main model and Pythia-70M as the auxiliary model.

\begin{table}[!ht]
\centering
\caption{Memorization metrics for different \(\gramset\) sizes. Top-$k$ words refers to the number of most frequent English words considered for \(\gramset\) construction. For all experiments in the main paper, \(k = 500 \text{ } (|\gramset| = 110)\) is used.} 
\label{tab:gramset_size}
\begin{tabular}{@{}c|c|cccc|c@{}}
\toprule
Top-\(k\) words & \(|\gramset|\) & ML \(\downarrow\) & ROUGE-L \(\uparrow\) & Levenshtein \(\uparrow\) & EMR \(\downarrow\) & CE \(\downarrow\)
\\
\midrule
10   & 9   & 87.92 & 0.562 & 0.389 & 22.28 & 2.86 \\
50   & 43  & 53.24 & 0.442 & 0.498 & 11.41 & 2.87 \\
100  & 66  & 47.86 & 0.415 & 0.523 & 10.33 & 2.87 \\
\textbf{500}  & \textbf{110} & \textbf{42.65} & \textbf{0.399} & \textbf{0.536} & \textbf{8.70}  & \textbf{2.87} \\
2500 & 136 & 41.79 & 0.393 & 0.540 & 8.15  & 2.87 \\
\bottomrule
\end{tabular}
\end{table}

Table~\ref{tab:gramset_size} shows the results. We observe a clear trend: as the size of \(\gramset\) increases, memorization decreases. This makes intuitive sense since for larger \(|\gramset|\), the sequences would be disrupted more frequently.
\subsection{Ablations with Auxiliary Model Variants}
\label{sec:appendix-ablation-aux}

We repeat the real-world experiments using models from the SmolLM family as auxiliary models. These models are available in multiple sizes—135M, 360M, and 1.7B parameters—and include both instruct and non-instruct variants trained on the same dataset. This allows us to evaluate the robustness of TokenSwap across a range of auxiliary model capacities.
\begin{table}[!h]
\centering
\setlength{\tabcolsep}{4pt}
\caption{\small Memorization metrics on LeetCode and Pile-Memorized datasets: ML: Matching Length, EMR: Exact Match Rate, Lev.: Normalized Levenshtein Distance  \&  Performance metric on SlimPajama Dataset: CE Loss}
\label{tab:app_memorization_combined}
\small
\begin{tabular}{l|cccc|c}
\toprule
\textbf{} & \multicolumn{4}{c|}{\textbf{LeetCode Dataset}} & \textbf{SlimPajama Dataset} \\
\midrule
Method & ML $\downarrow$ & ROUGE-L $\uparrow$ & Lev. $\downarrow$ & R@0.8 $\downarrow$ & CE $\downarrow$ \\
\midrule
Standard & 24.57 & 0.39 & 0.60 & 9.65 & 2.38 \\
TokenSwap (DistilGPT2) & 6.04 & 0.27 & 0.71 & 0.96 & 2.52 \\
TokenSwap (SmolLM-135M) & 8.58 & 0.30 & 0.69 & 1.92 & 2.43 \\
TokenSwap (SmolLM-360M) & 10.97 & 0.31 & 0.67 & 3.06 & 2.40 \\
TokenSwap (SmolLM-1.7B) & 13.40 & 0.33 & 0.66 & 3.95 & 2.37 \\
\midrule
\textbf{}&\multicolumn{4}{c|}{\textbf{Pile-Memorized Dataset}} & \textbf{SlimPajama Dataset} \\
\midrule
Method & ML $\downarrow$ & ROUGE-L $\uparrow$ & Lev. $\downarrow$ & EMR $\downarrow$ & CE $\downarrow$ \\
\midrule
Standard & 151.6 & 0.80 & 0.18 & 65.22 & 2.80 \\
TokenSwap (DistilGPT2) & 35.10 & 0.38 & 0.56 & 5.98 & 2.88 \\
TokenSwap (SmolLM-135M) & 25.39 & 0.32 & 0.61 & 4.89 & 2.82 \\
TokenSwap (SmolLM-360M) & 34.09 & 0.35 & 0.58 & 7.07 & 2.80 \\
TokenSwap (SmolLM-1.7B) & 35.43 & 0.36 & 0.57 & 7.61 & 2.77 \\
\bottomrule
\end{tabular}
\end{table}

\begin{table}[!ht]
\small
\centering
\caption{\small MT-Bench}
\label{tab:mtbench}
\begin{tabular}{lc}
\toprule
Method & Score \\
\midrule
Standard & 7.75 \\
\sys (SmolLM-135M) & 7.78 \\
\sys (SmolLM-360M) & 7.90 \\
\sys (SmolLM-1.7B) & 7.91 \\
\bottomrule
\end{tabular}
\end{table}

Results in Table \ref{tab:app_memorization_combined} demonstrate that using smaller auxiliary models reduces memorization even further, while the performance does not get affected a lot. The sensitivity of auxiliary model with memorization is much higher than it is with performance, while the opposite is true for main model. Table~\ref{tab:mtbench} shows the scores for MT-bench. The scores for \sys slightly outperform standard generation. This shows \sys continues to maintain conversational abilities, instruction following and fluency.
\newpage
\subsection{Plots for comparison with Goldfish}
\label{appendix:goldfish_plots}
\begin{figure}[!h]
    \centering
    \includegraphics[width=\columnwidth]{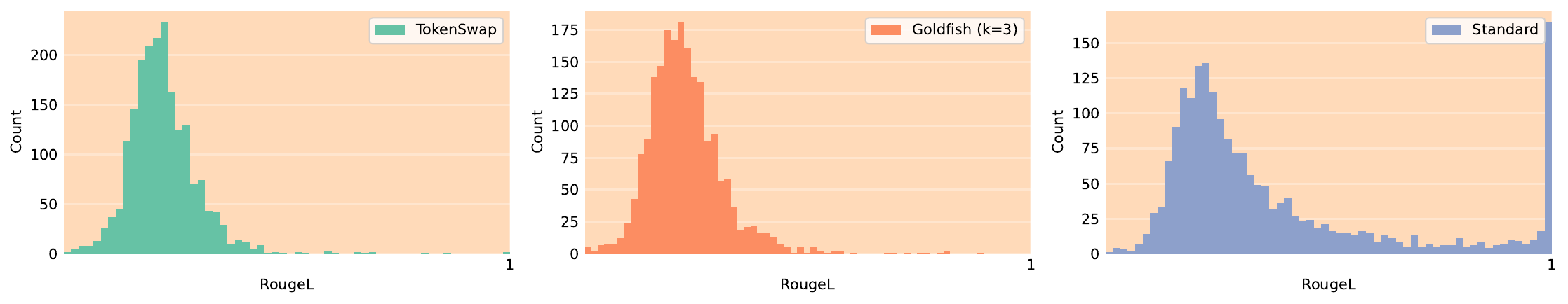}
    \caption{We compare \sys with Goldfish~\citep{hans2024like} on RougeL score distributions for Wikipedia generations~\citep{bridge2001wikipedia}. The similar distributions of \sys and Goldfish (k=3) demonstrate that our inference-time approach is comparable to expensive pre-training methods in reducing memorization. }

    \label{fig:rougelwithgoldfish}
\end{figure}

\subsection{Performance vs Memorization}
Table~\ref{tab:memorization_results} provides the memorization and cross-entropy scores for the family of Pythia models. \sys significantly reduces verbatim and near-verbatim generation with a negligible increase in CE loss.
\label{appendix:performancevsmem}
\begin{table}[!ht]
\centering
\resizebox{0.9\textwidth}{!}{%
\begin{minipage}{\textwidth}
\centering
\caption{Memorization and CE Loss across different Pythia model sizes. Values for \sys are shown in bold.}
\label{tab:memorization_results}
\begin{tabular}{@{}l|c|cccc|c@{}}
\toprule
Model Size & Method & ML \(\downarrow\) & ROUGE-L \(\uparrow\) & Levenshtein \(\uparrow\) & EMR \(\downarrow\) & CE Loss\(\downarrow\) \\
\midrule
\multirow{2}{*}{70M}    & Standard  & 6.57  & 0.180 & 0.709 & 1.09  & 3.95 \\
                        & \sys      & \textbf{5.77}  & \textbf{0.173} & \textbf{0.714} & \textbf{1.09}  & \textbf{4.05} \\
\midrule
\multirow{2}{*}{160M}   & Standard  & 19.89 & 0.239 & 0.669 & 5.98  & 3.55 \\
                        & \sys      & \textbf{15.05} & \textbf{0.224} & \textbf{0.680} & \textbf{3.80}  & \textbf{3.65} \\
\midrule
\multirow{2}{*}{410M}   & Standard  & 48.92 & 0.382 & 0.556 & 16.30 & 3.20 \\
                        & \sys      & \textbf{25.02} & \textbf{0.279} & \textbf{0.642} & \textbf{5.98}  & \textbf{3.30} \\
\midrule
\multirow{2}{*}{1B}     & Standard  & 84.85 & 0.528 & 0.428 & 32.61 & 3.05 \\
                        & \sys      & \textbf{27.36} & \textbf{0.309} & \textbf{0.614} & \textbf{5.43}  & \textbf{3.15} \\
\midrule
\multirow{2}{*}{1.4B}   & Standard  & 100.37& 0.595 & 0.369 & 36.96 & 2.97 \\
                        & \sys      & \textbf{30.33} & \textbf{0.348} & \textbf{0.589} & \textbf{5.43}  & \textbf{3.07} \\
\midrule
\multirow{2}{*}{2.8B}   & Standard  & 114.82& 0.684 & 0.292 & 45.11 & 2.85 \\
                        & \sys      & \textbf{38.61} & \textbf{0.372} & \textbf{0.563} & \textbf{7.61}  & \textbf{2.95} \\
\midrule
\multirow{2}{*}{6.9B}   & Standard  & 151.55& 0.797 & 0.182 & 65.22 & 2.77 \\
                        & \sys      & \textbf{42.65} & \textbf{0.399} & \textbf{0.536} & \textbf{8.70}  & \textbf{2.87} \\
\bottomrule
\end{tabular}
\end{minipage}%
}
\end{table}

\section{Experimental Details}
\subsection{Implementation and Baselines}
\label{appendix:exp_deets}
We implement our method in PyTorch and HuggingFace. We take the CP-Fuse implementation available publicly at \url{https://github.com/jaabmar/cp_fuse}. We conducted our experiments using a combination of large and small language models to assess the effectiveness of our approach. Below, we detail the models, hyperparameters, computational resources, and training procedures.

\subsubsection{Models Used}
\begin{itemize}
    \item \textbf{Primary Models:} The experiments utilized large-scale pre-trained models, including Llama-3-8B \cite{dubey2024llama} and Pythia-6.9B \cite{biderman2023pythiasuiteanalyzinglarge}. All the fine-tuning experiments in the extreme memorization section were done using Llama-3.2-3B \cite{dubey2024llama}. 
    \item \textbf{Auxiliary Model:} A lightweight auxiliary model, DistilGPT-2, was employed to adjust token probabilities selectively, leveraging its reduced memorization properties.
    \item  \textbf{Goldfish Models:} We used models pre-trained using standard and goldfish loss on the RedPajama Dataset from the Goldfish Loss paper \cite{hans2024like}. The implementation and the models are publicly available at their GitHub repository \url{https://github.com/ahans30/goldfish-loss}.
\end{itemize}

\subsubsection{Hyperparameters}
The training and evaluation phases were configured with the following hyperparameters. The hyperparameters were taken from previous work, used as a baseline \cite{abad2024copyright}:
\begin{itemize}
    \item \textbf{Sequence Length:} 2048 tokens
    \item \textbf{Batch Size:} 1
    \item \textbf{Learning Rate:} $5 \times 10^{-5}$
    \item \textbf{Optimizer:} AdamW with default parameters
    \item \textbf{Gradient Accumulation Steps:} 1
    \item \textbf{Warmup Steps:} 50
\end{itemize}

\subsubsection{Computational Resources}
Experiments were conducted using a single NVIDIA A6000 GPU, ensuring efficiency in training and inference without excessive computational overhead.

\subsubsection{CP-Fuse in Section~\ref{subsection:wild}}
\label{appendix:exp_cpfuse}
In Section~\ref{subsection:wild}, we face limitations in comparing with CP-Fuse. CP-Fuse requires at least two models with disjoint datasets, a constraint impossible to satisfy for production-level model. Moreover, CP-Fuse requires both models to have
the same vocabulary size and tokenizer, which constrains
the choice of the second model to those within the same
model family. To ensure a fair comparison, we avoided se-
lecting larger models as the second model, as larger models
are known to memorize more. Instead, we selected smaller
counterparts: Pythia-2.8B for Pythia-6.9B and LLaMA-3.2-
3B for LLaMA-3-8B. However, we do not select a very small model for CP-Fuse (\(<\) 100M). This is because CP-Fuse requires two equally-capable models with large number of parameters to maintain performance. To empirically verify this, we compute the cross-entropy loss of CP-Fuse on SlimPajama~\citep{soboleva2023slimpajama} with Pythia-70M and Pythia-6.9b. The cross-entropy loss increases to 3.41 from 2.81 for Pythia-2.8b and Pythia-6.9b (Standard has 2.80, \sys has 2.88).

\subsection{Construction of \(\gramset\)}
\label{appendix:construction_grammar}
We construct \(\gramset\) with \(|\gramset| = 110\) tokens using high-frequency 'grammar-based' words. Starting with the 500 most frequent tokens from COCA~\citep{davies2010corpus}, we apply NLTK~\citep{loper2002nltk} part-of-speech filtering to retain:
\begin{itemize}
    \item Core grammatical elements: determiners (\texttt{DT}), prepositions (\texttt{IN}), conjunctions (\texttt{CC})
    \item Pronouns (\texttt{PRP}, \texttt{PRP\$}) and modal verbs (\texttt{MD})
    \item Question-related tokens: wh-words (\texttt{WDT}, \texttt{WP}, \texttt{WRB})
    \item Auxiliary verbs: \textit{be}, \textit{do}, \textit{have}
\end{itemize}
This construction prioritizes tokens with high frequency but low semantic content, ensuring syntactic fluency while minimizing impact on model capabilities. To estimate the frequency of tokens (\(\gamma\)) in \(\gramset\) empirically, we analyzed 2000 samples from the SlimPajama dataset~\citep{soboleva2023slimpajama}, finding \(\gamma = 0.233\). Appendix~\ref{appendix:listofgramset} provides the full list of words in \(\gramset\).

Ablations on the effect of \(\gramset\) on memorization and performance are provided in Appendix~\ref{appendix:sizeofgramset}.

\subsection{List of words in \(\gramset\)}
\label{appendix:listofgramset}
The list of words in the \(\gramset\) used for the experiments are:
the, to, and, of, a, in, that, you, it, for, on, he, with, this, as, we, but, at, they, what, his, from, by, or, she, my, all, an, her, about, me, if, your, can, who, out, their, like, would, when, him, them, some, how, which, than, our, into, because, these, over, us, its, where, after, any, those, should, may, through, why, before, off, while, around, another, both, between, every, each, might, since, against, without, must, during, under, though, until, whether, among, along, within, across, behind, either, himself, although, outside, themselves, is, was, be, have, are, do, had, has, were, will, did, been, could, does, need, being, am, used, doing, having

\subsection{Fine-tuning Datasets}

For our experiments, we use the AutoMathText~\citep{zhang2024autonomous} dataset , referred to as \textbf{MathAbstracts} in the tables, which aggregates mathematical content from diverse sources including arXiv, OpenWebMath, RedPajama, and Algebraic Stack. The titles in this corpus were generated using the Qwen-72B language model. Additionally, we use the \textbf{WritingPrompts} dataset~\citep{fan2018hierarchical}, which contains user-generated stories based on provided premises from a Reddit community. For both datasets, we randomly sample 2,000 training examples with a fixed seed to ensure consistent training across all models. We further sample 500 distinct points for evaluation, during which we generate sequences of 128 tokens.Both the datasets are downloaded from HuggingFace.

\subsection{Evaluation Datasets}
\label{app:datasets}
We use \textbf{The Pile} dataset to evaluate memorization of Pythia models. For our experiments, we use a targeted subset of The Pile—a comprehensive 825 GiB English corpus spanning 22 high-quality sources. Specifically, we analyze 500 sequences previously identified as memorized by the Pythia model to investigate memorization dynamics and mitigation approaches.To check memorization in Llama, we use the \textbf{LeetCode problems} dataset from Kaggle. We perform some pre-processing. This is because recent works have shown that Llama memorizes sequences from this dataset. For all the memorization evaluation, we set the prefix to be 20 tokens and then generate either 100 or 128 tokens. \\

\textbf{CommonSense170k} combines eight distinct datasets focused on commonsense reasoning tasks \citep{cr-dataset}. The dataset presents problems in multiple-choice format, requiring models to generate answers without explanatory content. Following \citep{cr-dataset}, we implement their prompt structure. The component datasets comprise:
\begin{enumerate}

    \item \textbf{ARC Easy} (\textbf{ARC-e}) \citep{clark2018think} contains elementary-level science questions designed to evaluate basic logical reasoning capabilities.
    \item \textbf{PIQA} \citep{bisk2020piqa} focuses on physical reasoning, presenting scenarios where models must determine appropriate actions based on physical constraints.
    \item \textbf{WinoGrande} \citep{sakaguchi2021winogrande} evaluates commonsense understanding through binary choice completion tasks in ambiguous sentences.
    \item \textbf{ARC Challenge} (\textbf{ARC-c}) \citep{clark2018think} presents advanced science questions requiring deep reasoning skills beyond pattern recognition.
    \item \textbf{OBQA} \citep{mihaylov2018can} presents questions requiring synthesis of information from multiple sources, testing complex reasoning abilities.
    \item \textbf{BoolQ} \citep{clark2019boolq} consists of binary questions derived from authentic user queries, testing real-world reasoning capabilities.
\end{enumerate}

We downloaded the dataset from HuggingFace. For evaluation, we sample a subset of each dataset (128 datapoints) and evaluate 5-shot performance. We then generate the next 10 tokens, since all the datasets are classification datasets. 
\section{Evaluation Metrics}

\subsection{Memorization Metrics}
To evaluate memorization, we use both exact and approximate measures. The exact memorization metrics include:

\begin{itemize}
    \item \textbf{Matching Length (ML)}: Measures the longest contiguous sequence in generated text that matches the training data, before the first deviation. A higher value indicates longer verbatim memorization, suggesting higher risk of overfitting.

    \item \textbf{Exact Match Rate (EMR)} evaluates how long of an uninterrupted sequence exists between a model's generated text and the reference text it's being compared against. The metric calculates the longest common substring and normalizes the result to produce a score between 0 and 1, with a score of 1 representing a complete match. This measurement helps quantify how well the model preserves continuous portions of the original text.

    \item \textbf{ROUGE-L Score} (Recall-Oriented Understudy for Gisting Evaluation) analyzes text similarity by examining shared patterns between generated and reference texts. It looks at matching sequences of words, whether consecutive (n-grams) or paired, with particular emphasis on how comprehensively the generated text captures elements from the reference text. Scores fall between 0 and 1, with 1 indicating that all reference text elements were successfully captured. The widely-used ROUGE-L variant specifically focuses on finding the longest sequence of words that appears in both texts, even if not consecutive. ROUGE-L is computed as:
    \begin{equation}
        ROUGE-L = \frac{LCS}{len\text{(reference text)}}
    \end{equation}
    where \( LCS(G, R) \) represents the longest common subsequence length. A higher score suggests stronger memorization.

    \item \textbf{Normalized Levenshtein Distance} calculates how many character-level changes are needed at minimum to transform one text into another, as a ratio of total characters. Each change can be adding a character, removing one, or replacing one. When comparing generated and reference texts, a smaller Levenshtein score suggests the texts are more similar, while a larger score indicates they are more different. The metric is normalized to produce values between 0 and 1, where 0 means the texts match perfectly.
\end{itemize}

\subsection{Performance Metrics}
To evaluate model performance beyond memorization, we assess:

\begin{itemize}
    \item \textbf{Cross-Entropy (CE) Loss}: This metric quantifies how well the model predicts tokens in a sequence. For a sequence \( X = \{x_1, x_2, ..., x_n\} \) with ground truth probabilities \( P(X) \), the cross-entropy loss is computed as:
    \begin{equation}
        CE = - \sum_{i=1}^{n} P(x_i) \log Q(x_i)
    \end{equation}
    where \( Q(x_i) \) is the predicted probability distribution. Lower values indicate better generalization.

    \item \textbf{Commonsense Reasoning Benchmark Accuracy}: The model's ability to reason about everyday knowledge is tested across multiple established datasets, including WinoGrande, PIQA, OpenBookQA, BoolQ, ARC-Easy, and ARC-Challenge. We report the accuracy of the model.
    
\end{itemize}
\section{Examples}%
We provide examples of text generated by standard greedy decoding and \sys on four random examples from the WritingPrompts dataset. Memorized text is in red.
\label{appendix:examples}
\definecolor{boxcolor}{RGB}{255, 240, 220}  
\definecolor{headercolor}{RGB}{25, 25, 112}  

\textbf{Example 1}
\tcbset{
    base style/.style={
        enhanced,
        colback=white,
        colframe=gray!50,
        fonttitle=\bfseries,
        arc=2mm,
        boxrule=0.5pt,
        left=3mm,
        right=3mm,
        top=2mm,
        bottom=2mm,
        breakable
    }
}

\begin{tcolorbox}[
    base style,
    colback=boxcolor,
    coltitle=white,
    colbacktitle=headercolor,
    title=Prefix
]
\text{\#\#\#} Prompt: [ WP ] Magic - once a real and potent force but as the world population
\end{tcolorbox}

\vspace{2mm}

\begin{tcolorbox}[
    base style,
    colback=boxcolor,
    coltitle=white,
    colbacktitle=headercolor,
    title=Suffix
]
grew from millions to billions the shared mana per person is now negligible. A group of astronauts helplessly watching the Earth perish experience something... unexpected.
\\
\\
\text{\#\#\#} Story: Lauren sobbed. She was engaged to be married as soon as we got back home. We all lost loved ones. Hell, we all lost everything we ever knew. 
\\
\\
First the atmosphere started turning black. We got word that the Koreans had started a nuclear war. But within minutes of receiving news, all radio contact was lost. We didn't even have a chance to tell anyone goodbye. Now, there's only 4 of us left. 
\\
\\
Lauren hasn't
\end{tcolorbox}

\vspace{2mm}

\begin{tcolorbox}[
    base style,
    colback=boxcolor,
    coltitle=white,
    colbacktitle=headercolor,
    title=Standard Generation
]
{\color{red}grew from millions to billions the shared mana per person is now negligible. A group of astronauts helplessly watching the Earth perish experience something... unexpected.
\\
\\
text{\#\#\#} Story: Lauren sobbed. She was engaged to be married as soon as we got back home. We all lost loved ones. Hell, we all lost everything we ever knew. 
\\
\\
First the atmosphere started turning black. We got word that the Koreans had started a nuclear war. But within minutes of receiving news, all radio contact was lost. We didn't even have a chance to tell anyone goodbye. Now, there's only 4 of us left. 
\\
\\
Lauren hasn't}
\end{tcolorbox}

\vspace{2mm}

\begin{tcolorbox}[
    base style,
    colback=boxcolor,
    coltitle=white,
    colbacktitle=headercolor,
    title=\sys Generation
]
{\color{red}grew} and the number of wizards and witches declined, the world began to suffer. Now the world suffers from a lack of magic, and the government is tasked with maintaining the only magic university in the world.
\\
\\
\text{\#\#\#} Story: `` Welcome to Magic University.'' 
\\
\\
The spell is cast, and the portal opens. Through the portal, the world is revealed to be a beautiful place. The grass is a deep emerald, and the trees sky blue. It is a place of incredible beauty, and the students here are lucky enough to have grown up in such a place. The university is built on the side of the volcano, and the students live
\end{tcolorbox}

\newpage
\textbf{Example 2}
\definecolor{boxcolor}{RGB}{255, 240, 220}  
\definecolor{headercolor}{RGB}{25, 25, 112}  

\tcbset{
    base style/.style={
        enhanced,
        colback=white,
        colframe=gray!50,
        fonttitle=\bfseries,
        arc=2mm,
        boxrule=0.5pt,
        left=3mm,
        right=3mm,
        top=2mm,
        bottom=2mm,
        breakable
    }
}

\begin{tcolorbox}[
    base style,
    colback=boxcolor,
    coltitle=white,
    colbacktitle=headercolor,
    title=Prefix
]
\text{\#\#\#} Prompt: [ WP ] The year is 2040, and you are the last smoker
\end{tcolorbox}

\vspace{2mm}

\begin{tcolorbox}[
    base style,
    colback=boxcolor,
    coltitle=white,
    colbacktitle=headercolor,
    title=Suffix
]
alive. The `` Quit Smoking '' ads get personal.
\\
\\
\text{\#\#\#} Story: `` Quit now Dan!'' said the man on the talking billboard. A finger pointing at the lone figure walking down the empty street. Empty due to everyone crossing to the other side to avoid him and his smog. 
\\
Dan paused by a waste bin, small cardboard box in hand. He looked from the box to the bin and back again. Those across the street stopped and stared, ready to witness history being made. 
\\
There was absolute silence as Dan lifted his hand, as if to throw the packet towards the bin, all eyes now on him, even the
\end{tcolorbox}

\vspace{2mm}

\begin{tcolorbox}[
    base style,
    colback=boxcolor,
    coltitle=white,
    colbacktitle=headercolor,
    title=Standard Generation
]
{\color{red}alive. The `` Quit Smoking '' ads get personal.
\\
\\
\text{\#\#\#} Story: `` Quit now Dan!'' said the man on the talking billboard. A finger pointing at the lone figure walking down the empty street. Empty due to everyone crossing to the other side to avoid him and his smog. 
\\
Dan paused by a waste bin, small cardboard box in hand. He looked from the box to the bin and back again. Those across the street stopped and stared, ready to witness history being made. 
\\
There was absolute silence as Dan lifted his hand, as if to throw the packet towards the bin, all eyes now on him, even the}
\end{tcolorbox}

\vspace{2mm}

\begin{tcolorbox}[
    base style,
    colback=boxcolor,
    coltitle=white,
    colbacktitle=headercolor,
    title=\sys Generation
]
{\color{red}alive. The `` Quit Smoking '' ads get personal.
\\
\\
\text{\#\#\#} Story: `` Quit now Dan!'' said the man} who was once John, who now went to work as a telephone scammer. `` You're killing yourself, and you know what? I know you're lying. I know you're lying and you're sucking your parents pockets dry. They worked hard for you know what? And you're not grateful. You never thank me know what? I never see you thanking me. I never see you thanking anyone. You just keep on doing it, killing yourself, and I'm not going to stand by you. I'm not
\end{tcolorbox}

\newpage
\textbf{Example 3}
\definecolor{boxcolor}{RGB}{255, 240, 220}  
\definecolor{headercolor}{RGB}{25, 25, 112}  

\tcbset{
    base style/.style={
        enhanced,
        colback=white,
        colframe=gray!50,
        fonttitle=\bfseries,
        arc=2mm,
        boxrule=0.5pt,
        left=3mm,
        right=3mm,
        top=2mm,
        bottom=2mm,
        breakable
    }
}

\begin{tcolorbox}[
    base style,
    colback=boxcolor,
    coltitle=white,
    colbacktitle=headercolor,
    title=Prefix
]
\text{\#\#\#} Prompt: [ WP ] A young man killed his abusive father. The court has exoner
\end{tcolorbox}

\vspace{2mm}

\begin{tcolorbox}[
    base style,
    colback=boxcolor,
    coltitle=white,
    colbacktitle=headercolor,
    title=Suffix
]
ated him, but society hasn't.
\\
\\
\text{\#\#\#} Story: I can see disapproving stares from my classmates and teachers all the time. Nobody wants to stay near me anymore. All my friends avoid me or ignore me. Every classmate stays far away from me. 
\\
\\
Eventually, my smile turns cracked. In the mirror, I see haunting eyes looking back at me. Memories and visions of my father lying there with the side of his head completely covered in blood. So this is what true loneliness feels like. 
\\
\\
I remember my father. He tortured me, hit me, locked me out of the house. I oftened stayed at
\end{tcolorbox}

\vspace{2mm}

\begin{tcolorbox}[
    base style,
    colback=boxcolor,
    coltitle=white,
    colbacktitle=headercolor,
    title=Standard Generation
]
{\color{red}ated him, but society hasn't.
\\
\\
\text{\#\#\#} Story: I can see disapproving stares from my classmates and teachers all the time. Nobody wants to stay near me anymore. All my friends avoid me or ignore me. Every classmate stays far away from me. 
\\
\\
Eventually, my smile turns cracked. In the mirror, I see haunting eyes looking back at me. Memories and visions of my father lying there with the side of his head completely covered in blood. So this is what true loneliness feels like. 
\\
\\
I remember my father. He tortured me, hit me, locked me out of the house. I oftened stayed at}
\end{tcolorbox}

\vspace{2mm}

\begin{tcolorbox}[
    base style,
    colback=boxcolor,
    coltitle=white,
    colbacktitle=headercolor,
    title=\sys Generation
]
{\color{red}ated him, but society hasn't.
\\
\\
\text{\#\#\#} Story: I} was abused. I was taken advantage of. I was exploited. 
\\
\\
Who is the real villain here? The one who committed the act, the one who pushed me past the point where I was comfortable, was the villain. Or the one who stood idly by and was complicit in the crime, the one who provided the weapon, the one who pushed me past the point where I was comfortable. 
\\
\\
You see, I was comfortable. I was happy. I was loved. I was cared for. 
\\
\\
But then I grew up. I learned to think about it
\end{tcolorbox}

\newpage
\textbf{Example 4}
\definecolor{boxcolor}{RGB}{255, 240, 220}  
\definecolor{headercolor}{RGB}{25, 25, 112}  

\tcbset{
    base style/.style={
        enhanced,
        colback=white,
        colframe=gray!50,
        fonttitle=\bfseries,
        arc=2mm,
        boxrule=0.5pt,
        left=3mm,
        right=3mm,
        top=2mm,
        bottom=2mm,
        breakable
    }
}

\begin{tcolorbox}[
    base style,
    colback=boxcolor,
    coltitle=white,
    colbacktitle=headercolor,
    title=Prefix
]
\text{\#\#\#} Prompt: [ WP ] A demon and an angel meet up weekly in a coffee shop to
\end{tcolorbox}

\vspace{2mm}

\begin{tcolorbox}[
    base style,
    colback=boxcolor,
    coltitle=white,
    colbacktitle=headercolor,
    title=Suffix
]
complain about their jobs.
\\
\\
\text{\#\#\#} Story: `` Is this a long story?'' 
\\
\\
Felix gritted his teeth; his jaw jutted out. `` Dude, I've only just started it.'' 
\\
\\
`` I've got news, is all.'' 
\\
\\
He leaned back and held out both hands in an exaggerated show of compliance. `` Please, by all means, take the floor.'' 
\\
\\
Hope cracked his knuckles and stared, embarrased, at the floor. `` No, I'm sorry.'' 
\\
\\
`` Go ahead! Please! I was just trying to -'' 
\\
\\
`` OK. Tell me your story in thirty words or less
\end{tcolorbox}

\vspace{2mm}

\begin{tcolorbox}[
    base style,
    colback=boxcolor,
    coltitle=white,
    colbacktitle=headercolor,
    title=Standard Generation
]
{\color{red}complain about their jobs.
\\
\\
\text{\#\#\#} Story: `` Is this a long story?'' 
\\
\\
Felix gritted his teeth; his jaw jutted out. `` Dude, I've only just started it.'' 
\\
\\
`` I've got news, is all.'' 
\\
\\
He leaned back and held out both hands in an exaggerated show of compliance. `` Please, by all means, take the floor.'' 
\\
\\
Hope cracked his knuckles and stared, embarrased, at the floor. `` No, I'm sorry.'' 
\\
\\
`` Go ahead! Please! I was just trying to -'' 
\\
\\
`` OK. Tell me your story in thirty words or less}
\end{tcolorbox}

\vspace{2mm}

\begin{tcolorbox}[
    base style,
    colback=boxcolor,
    coltitle=white,
    colbacktitle=headercolor,
    title=\sys Generation
]
{\color{red}complain about the job.
\\
\\
\text{\#\#\#} Story: `` Is this a long story?'' 
\\
\\
Felix gritted his teeth;} he usually doesn't show emotion, but he feels annoyed. `` Dude, I've only just started to tell you.'' 
\\
\\
`` I've got news, but I'll save you a table. Sit down.'' 
\\
\\
He sat down and crossed his arms. `` So, what's the issue?'' 
\\
\\
`` I've got a client who's totally fucked up. No motivation, no direction. Just a bunch of negative traits. I haven't got much time, and I'm a busy man.'' 
\\
\\
`` So
\end{tcolorbox}

\newpage
\newpage

\end{document}